\documentclass{article}

\usepackage[table,xcdraw]{xcolor} 

\usepackage{microtype}
\usepackage{inconsolata} 
\usepackage{graphicx}
\usepackage{booktabs} 
\usepackage{silence}
\usepackage{hyperref}
\usepackage{xr-hyper}

\usepackage[accepted]{mlsys2025} 

\usepackage[T1]{fontenc} 
\usepackage[utf8]{inputenc} 
\usepackage{multirow} 
\usepackage{tabularx} 
\usepackage{color} 
\usepackage{textcomp} 
\usepackage{tipa}
\usepackage{amsmath} 
\usepackage{amssymb} 
\usepackage{amsxtra} 
\usepackage{wasysym} 
\usepackage{isomath} 
\usepackage{mathtools} 
\usepackage{txfonts} 
\usepackage{upgreek} 
\usepackage{enumerate} 
\usepackage{tensor} 
\usepackage{pifont} 
\usepackage{ulem} 
\usepackage{xfrac} 
\usepackage{arydshln}
\usepackage{float}
\usepackage[skip=5pt plus1pt, indent=0pt]{parskip}
\definecolor{color-E8E8E8}{rgb}{0.91,0.91,0.91}
\definecolor{color-1F1F1F}{rgb}{0.12,0.12,0.12}
\definecolor{color-2B2B2B}{rgb}{0.17,0.17,0.17}
\definecolor{color-0E2841}{rgb}{0.05,0.16,0.25}
\definecolor{color-0F4761}{rgb}{0.06,0.28,0.38}
\definecolor{color-C1E4F5}{rgb}{0.76,0.89,0.96}
\definecolor{color-CC7832}{rgb}{0.8,0.47,0.2}
\definecolor{color-A9B7C6}{rgb}{0.66,0.72,0.78}
\definecolor{color-AA4926}{rgb}{0.67,0.29,0.15}
\definecolor{color-6A8759}{rgb}{0.42,0.53,0.35}
\definecolor{color-8888C6}{rgb}{0.53,0.53,0.78}
\definecolor{codebg}{gray}{0.98}
\definecolor{codeframe}{gray}{0.85}
\definecolor{codenums}{gray}{0.55}
\definecolor{codecm}{gray}{0.40}

\usepackage{listings}
\usepackage{xcolor}
\usepackage{tablefootnote}

\usepackage{longtable}

\definecolor{codegreen}{rgb}{0,0.6,0}
\definecolor{codegray}{rgb}{0.5,0.5,0.5}
\definecolor{codepurple}{rgb}{0.58,0,0.82}
\definecolor{backcolour}{rgb}{0.95,0.95,0.92}

\lstdefinestyle{pub-bw}{
  language=Python,
  backgroundcolor=\color{codebg},
  basicstyle=\ttfamily\fontsize{9}{10.5}\selectfont\linespread{0.97},
  keywordstyle=\bfseries,
  stringstyle=\itshape,
  commentstyle=\itshape\color{codecm},
  numbers=left,
  numberstyle=\scriptsize\color{codenums},
  numbersep=8pt,
  frame=single,
  rulecolor=\color{codeframe},
  framerule=0.4pt,
  framesep=6pt,
  xleftmargin=0em,
  framexleftmargin=0em,
  breaklines=true,
  breakatwhitespace=true,
  columns=fullflexible,
  keepspaces=true,
  upquote=true,
  showstringspaces=false,
  tabsize=2,
  captionpos=b,
  aboveskip=1.0em,
  belowskip=1.0em,
  numbers=none, 
}

\usepackage{xurl} 
\usepackage{url}            
\usepackage{amsfonts}       
\usepackage{nicefrac}       
\usepackage{dsfont}
\usepackage{amssymb}

\usepackage{caption}
\usepackage{subcaption}

\usepackage{svg}

\usepackage{enumitem}
\setlist[itemize]{noitemsep, topsep=0pt, parsep=0pt, partopsep=0pt}
\setlist[enumerate]{noitemsep, topsep=0pt, parsep=0pt, partopsep=0pt}

\usepackage[compact]{titlesec}
\titlespacing*{\section}{0pt}{0.6ex plus .1ex minus .1ex}{0.4ex plus .1ex}
\titlespacing*{\subsection}{0pt}{0.4ex plus .1ex minus .1ex}{0.2ex plus .1ex}
\titlespacing*{\subsubsection}{0pt}{0.3ex plus .1ex minus .1ex}{0.1ex plus .1ex}
\setcitestyle{numbers,square}

\usepackage{tabularx,booktabs,array}
\renewcommand{\arraystretch}{1.15} 
\newcolumntype{P}[1]{>{\raggedright\arraybackslash}p{#1}} 
\newcolumntype{Y}{>{\raggedright\arraybackslash}X}        

\mlsystitlerunning{Agent Spec}

\begin{document}

\twocolumn[
\mlsystitle{Open Agent Specification (Agent Spec): A Unified Representation for AI Agents}




\begin{mlsysauthorlist}
\mlsysauthor{Soufiane Amini}{oracle}
\mlsysauthor{Yassine Benajiba}{oracle}
\mlsysauthor{Cesare Bernardis}{oracle}
\mlsysauthor{Paul Cayet}{oracle}
\mlsysauthor{Hassan Chafi}{oracle}
\mlsysauthor{Abderrahim Fathan}{oracle}
\mlsysauthor{Louis Faucon}{oracle}
\mlsysauthor{Damien Hilloulin}{oracle}
\mlsysauthor{Sungpack Hong}{oracle}
\mlsysauthor{Ingo Kossyk}{oracle}
\mlsysauthor{Tran Minh Son Le}{oracle}
\mlsysauthor{Rhicheek Patra}{oracle}
\mlsysauthor{Sujith Ravi}{oracle}
\mlsysauthor{Jonas Schweizer}{oracle}
\mlsysauthor{Jyotika Singh}{oracle}
\mlsysauthor{Shailender Singh}{oracle}
\mlsysauthor{Weiyi Sun}{oracle}
\mlsysauthor{Kartik Talamadupula}{oracle}
\mlsysauthor{Jerry Xu}{oracle}
\end{mlsysauthorlist}

\mlsysaffiliation{oracle}{Oracle Corporation}

\mlsyscorrespondingauthor{Rhicheek Patra}{rhicheek.patra@oracle.com}

\mlsyskeywords{Machine Learning, MLSys}

\vskip 0.3in

\begin{abstract}
\vspace*{15pt}
The proliferation of agent frameworks has led to fragmentation in how agents are defined, executed, and evaluated. Existing systems differ in their abstractions, data flow semantics, and tool integrations, making it difficult to share or reproduce workflows. We introduce \textbf{Open Agent Specification (Agent Spec)}, a declarative language that defines AI agents and agentic workflows in a way that is compatible across frameworks, promoting reusability, portability and interoperability of AI agents. Agent Spec defines a common set of components, control and data flow semantics, and schemas that allow an agent to be defined once and executed across different runtimes. Agent Spec also introduces a standardized \textbf{Evaluation harness} to assess agent behavior and agentic workflows across runtimes - analogous to how HELM and related harnesses standardized LLM evaluation - so that performance, robustness, and efficiency can be compared consistently across frameworks. We demonstrate this using \textbf{four distinct runtimes} (LangGraph, CrewAI, AutoGen, and WayFlow) evaluated over \textbf{three different benchmarks} (\texttt{SimpleQA Verified}, \texttt{$\tau^2$-Bench} and \texttt{BIRD-SQL}). We provide accompanying toolsets: a Python SDK (PyAgentSpec), a reference runtime (WayFlow), and adapters for popular frameworks (e.g., LangGraph, AutoGen, CrewAI). Agent Spec bridges the gap between model-centric and agent-centric standardization \& evaluation, laying the groundwork for reliable, reusable, and portable agentic systems.

\end{abstract}
]



\printAffiliationsAndNotice{Authors listed alphabetically.}  

\section{Introduction}
\label{ref-002}

The rapid emergence of AI agent frameworks has accelerated the development of systems capable of planning, invoking tools, and coordinating multi-step workflows \cite{SAPKOTA2026103599, orclblog, xie2025ai, wolflein-etal-2025-llm}. However, this progress has also introduced fragmentation: each framework defines its own abstractions, configuration formats, and execution assumptions \cite{derouiche2025agenticaiframeworksarchitectures, kapoor2025ai}. As a result, organizations struggle when porting agents across stacks, evaluating behaviors consistently, or reusing designs developed by other teams. These challenges hinder reproducibility, slow down prototyping-to-deployment, and complicate governance in enterprise environments \cite{10.1145/3711896.3736570, chen2025aiopslab}.

Concurrently, several standardization efforts are shaping the agent ecosystem. Protocols such as Anthropic’s Model Context Protocol (MCP)~\cite{mcp} focus on resource and tool provisioning; inter-agent communication initiatives like Google’s Agent2Agent~\cite{a2a} and BeeAI’s Agent Communication Protocol (ACP)~\cite{acp} standardize message exchange. In addition, the Agntcy initiative~\cite{agntcy} aims to promote open interoperability and governance standards for the emerging agent ecosystem, providing shared principles for agent definition, discovery, and interaction across frameworks. While these efforts advance interoperability at the resource and communication layers, there still exists a critical gap at the level of agent behavior and execution semantics. In practice, this means there is no common standard for how control and data flows are managed, how agent state evolves, or how inputs and outputs are validated, leading to inconsistencies in agent execution behavior across different runtime environments.

To address these gaps, we introduce \textbf{Agent Spec}, a declarative, framework-agnostic configuration language for defining AI agents and their workflows with high fidelity. By providing a unified representation, Agent Spec enables seamless reusability, portability and interoperability across diverse frameworks, ensuring consistent behavior and reproducibility. 

In practice, Agent Spec serves as an abstraction layer above framework-specific implementations, providing a unifying interface that captures agent functionality beyond individual framework constraints. It makes evaluation a first-class concern by defining common execution semantics and measurement protocols, so that agent and workflow performance can be compared across frameworks - similar in spirit to how HELM~\cite{liang2023holistic} and related harnesses standardized LLM evaluation. Like HELM, EleutherAI's \texttt{lm-evaluation-harness}~\cite{eleutherai2024lmEvalHarness} and \texttt{MT-Bench}~\cite{bai2024mtbench101}, Agent Spec centers evaluation as a core principle. However, while these efforts focus on model-centric standardized assessment, Agent Spec standardizes agent- and workflow-centric standardized assessment across frameworks. By fixing a common intermediate representation, execution semantics, and measurement protocol, Agent Spec supports reproducible, cross-platform comparisons of agent quality, safety, robustness, and efficiency, thus bridging the gap between agent portability, reusability and rigorous evaluation.

Our contributions in this work are as follows: 

\begin{itemize}
    \item We present \textbf{Agent Spec}, a declarative, framework-agnostic language for specifying agents and workflows with high fidelity, enabling define-once, run-anywhere portability and seamless reusability across diverse runtimes.
    \item We formalize a core set of components and execution semantics - \textbf{Agents}, \textbf{Flows}, \textbf{Nodes}, \textbf{Tools}, and \textbf{control/data flow edges} and discuss design principles for reliable, composable, and reusable agentic systems.
    \item We release \textbf{accompanying toolsets}: a Python SDK (\texttt{PyAgentSpec}~\footnote{\url{https://pypi.org/project/pyagentspec/}}) for programmatic authoring and serialization, a reference runtime (\texttt{WayFlow}~\footnote{\url{https://github.com/oracle/wayflow}}) with native Agent Spec support, and runtime adapters for popular frameworks to demonstrate cross-framework execution.
    \item Agent Spec serves as an \textbf{Evaluation harness}, enabling consistent comparison of agent and flow patterns across different frameworks. Using this capability, we report initial results on representative tasks from three benchmark datasets. Interestingly, we observe that both LangGraph and WayFlow deliver reliable accuracy and query time stability across datasets, AutoGen outperforms on complex reasoning tasks (e.g., on $\tau^2$-Bench~\cite{barres2025tau2}) but with increased latency, and CrewAI's multi-agent coordination yields limited accuracy gains despite longer query times.
\end{itemize}

\section{Background \& Motivation}
\label{ref-003}

\subsection{Related Work}
\label{ref-related-work}

AI agent frameworks provide the building blocks for constructing agentic systems that can reason, make decisions, and complete multi-step tasks. In recent years, with the advent of large language models (LLMs), several general-purpose frameworks have emerged with distinct abstractions, focus areas, and strengths, including \textbf{LangChain}~\cite{LangChain}, \textbf{LangGraph}~\cite{LangGraph}, \textbf{GPT-Functions}~\cite{openaiFunctionCalling}, \textbf{AutoGen}~\cite{autogen}, \textbf{OCI Agents}~\cite{ociagents}, \textbf{Semantic Kernel}~\cite{semanticKernel}, and \textbf{CrewAI}~\cite{crewAI}. These frameworks can augment LLMs (e.g., through Agents, Flows) with structured planning~\cite{wang2022self}, memory~\cite{wang2023augmenting}, tool use~\cite{schick2023toolformer}, and the ability to perform search and access external environments~\cite{zhu2023large, kim2024rada}. 

LangChain~\cite{LangChain} offers a flexible, modular SDK for composing applications with LLMs, tools, long-term memory, and retrieval, and it underpins a large ecosystem of integrations. LangGraph~\cite{LangGraph} builds on LangChain to model agents and workflows as stateful, directed graphs with explicit control flow, enabling deterministic branching, loops, and multi-actor coordination; its associated tooling (e.g., LangGraph Studio) provides visualization and debugging support. AutoGen~\cite{autogen} focuses on multi-agent conversations with tool use and human-in-the-loop capabilities, providing abstractions for autonomous agent roles, inter-agent messaging, and extensible tool execution, which makes it suitable for complex, iterative workflows. It is designed with an open and flexible conversational communication style between agents, and its event-driven and distributed architecture makes it suitable for workflows that require long-running agents that collaborate across information boundaries with variable degrees of human involvement.
Semantic Kernel~\cite{semanticKernel} is a model-agnostic SDK (with support across multiple programming languages) that provides planners, memories, connectors, and orchestration utilities to integrate LLMs and data sources into production applications. OCI AI Agent Platform~\cite{ociagents} targets enterprise-grade deployments with built-in tools and patterns (e.g., natural language SQL generation, retrieval-augmented generation), emphasizing scalability, manageability, and integration with enterprise services. CrewAI~\cite{crewAI} centers on multi-agent collaboration with explicit roles, tasks, and shared state, offering opinionated primitives for coordination and memory.

Separately, GPT-Functions (tool/function calling) \cite{openaiFunctionCalling} specifies how models such as GPT-4o can invoke user-defined tools through structured function signatures, enabling safer and more controllable LLM-tool interaction across many runtimes. OctoTools \cite{lu2025octotools} proposes an extensible framework oriented toward complex reasoning with ``tool cards'' that standardize wrappers around heterogeneous tools to improve transparency, reduce integration errors, and simplify maintenance.

While these frameworks advance agentic capabilities in complementary directions - ranging from SDKs to orchestration, coordination, and tool interoperability - the ecosystem remains fragmented. Differences in configuration models, component semantics, and execution assumptions hinder reusability, reliability, and reproducibility. This fragmentation motivates a unifying, declarative, and framework-agnostic specification such as \textbf{Agent Spec}, which provides a reusable \& portable representation of agents and workflows executable across diverse runtimes.

\subsection{Motivation} 
\label{ref-motivation}

Existing agentic frameworks have their own strengths and limitations (e.g., limited tool types, lack of generalization across diverse tasks) \cite{raheem2025agentic, huang2025ai, joshi2025review}. While flexibility is crucial to accommodating diverse agentic design patterns, currently available frameworks do not fully support all of them. \textbf{Table \ref{tab:AgentSpec-topic-table}} outlines key objectives that a comprehensive framework should cover and describes how Agent Spec addresses each of these requirements.

\begin{table*}[t]
\small
\centering
\caption{Addressing challenges and objectives in agentic frameworks through the Open Agent Specification.}
\begin{tabularx}{0.95\textwidth}{p{2.2cm}p{5.0cm}p{8.0cm}}
\toprule
\textbf{Topic} & \textbf{Objective} & \textbf{Open Agent Specification} \\
\midrule
Agents &
Author and edit conversational solutions such as ReAct \cite{yao2023react} style agents. &
Define “resource” components (LLMs, memory, tools, or agents) attachable to other components’ properties. \\
\midrule
Flows &
Author and edit graph-style workflows such as multi-step business processes. &
Define components with well-defined execution paths (such as directed graphs), supporting conditional or looping routes. \\
\midrule
Control flow routing &
Ability to define the flow of execution between components in the system, including conditional or looping flows. &
Define control flow edges enabling one-to-many or many-to-one transitions where the specific transition to be taken is determined at runtime for a given execution. \\
\midrule
I/O routing &
Ability to define how outputs feed inputs for fixed, conditional, or looping flows, including reuse across steps. &
Define data flow edges supporting one-to-many or many-to-one transitions where the specific transition to be taken is determined at runtime for a given execution.
\\
\midrule
Nesting \& Composition &
Reuse components or solutions at any level of granularity in other solutions; includes multi-agent composition. &
Define components that can be encapsulated and reused; control message visibility, and orchestration of execution order and dependency either statically or dynamically.\\
\midrule
Referencing &
Ability to define a component instance once and refer to it in multiple places. &
Use reference syntax to specify property values and reuse defined components across multiple flows or tools. For example, define an Agent once and reference it via an AgentNode in a Flow.\\
\bottomrule
\end{tabularx}
\label{tab:AgentSpec-topic-table}
\end{table*}

Moreover, different frameworks are usually parameterized and configured by different means. This makes porting solutions between frameworks a tedious and error-prone process, hindering collaboration and knowledge sharing among teams developing agentic applications. 

To address these challenges, a declarative, shareable, portable, and framework-independent representation of agentic design is essential. Analogous to how ONNX~\cite{onnx} standardized ML model interchange across frameworks such as PyTorch~\cite{pytorch} and TensorFlow~\cite{tensorflow}, \textbf{Agent Spec} aims to define a common interchange format for AI agents and workflows. Just as ONNX enables models trained in one framework to be executed in another without reimplementation, Agent Spec allows agents to be authored once and deployed consistently across multiple frameworks.

\subsection{Agent Spec within the Ecosystem} 
\label{ref-028}

Agent Spec streamlines the design of agentic assistants and workflows by providing a unified, framework-agnostic configuration language that defines agent patterns, components, and behaviors. It complements, rather than replacing, ongoing standardization efforts such as:

\begin{itemize}
\item \textbf{Anthropic's Model Context Protocol (MCP)} \cite{mcp}: Standardizes resource and data provision via client/server based RESTful APIs.

\item \textbf{Google's Agent2Agent Protocol} \cite{a2a} and \textbf{BeeAI's Agent Communication Protocol (ACP)} \cite{acp}: Standardized APIs for distributed agent communication.

\item The \textbf{Agntcy initiative}~\cite{agntcy} fosters open interoperability and governance standards, defining shared principles for agent discovery and cross-framework collaboration.
\end{itemize}

\begin{figure}[t]
\centering
\includegraphics[width=1\linewidth]{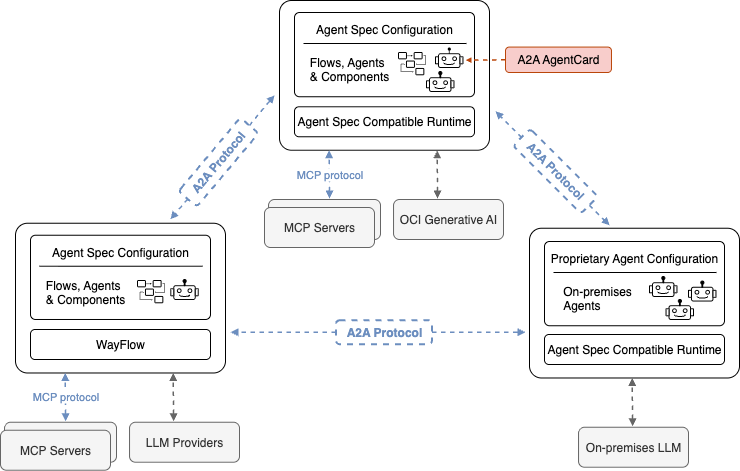} 
\caption{Position of Agent Spec within the emerging agent-standardization stack. MCP governs resource and data provisioning, A2A/ACP specify inter-agent communication, and Agent Spec defines the declarative layer for agent behavior and execution semantics.}
\label{fig:4}
\end{figure}

While protocols like MCP and A2A standardize tool or resource provisioning as well as inter-agent communication, Agent Spec complements these efforts by enabling standardized configuration of components related to agentic system design and execution in general, as shown in Fig. \ref{fig:4}.

\section{Agent Spec: Objectives and Contributions}
\subsection{Objectives of Open Agent Specification}
\label{ref-004}

Agent Spec defines a common representation that generalizes the capabilities for AI agents and workflows. Agent Spec graphs can be executed in different frameworks through runtimes that map its components to framework-specific primitives. Combined with importers and exporters for various frameworks, this enables transparent portability of agentic workflows. Agent Spec also provides validation mechanisms to ensure that workflows are syntactically and logically correct before execution, promoting consistency and reliability.

\subsection{Key Contributions of Agent Spec}
\label{ref-021}

Agent Spec offers several key contributions that address current challenges in the development and deployment of agentic workflows. The following subsections highlight how Agent Spec improves portability, reusability, interoperability, and unification across diverse frameworks, ultimately enhancing the effectiveness and robustness of AI agent systems.

\subsubsection{Portability \& Reusability}
\label{ref-022}


\textbf{Framework-Agnostic:} Agent Spec abstracts agent definitions from specific implementations by providing a unified declarative representation, allowing agents to operate uniformly across multiple platforms (e.g., AutoGen, LangGraph, WayFlow) without any need for reimplementation.

\textbf{Modular Design:} Its component-based architecture promotes reusability and extensibility, allowing complex agentic systems to be composed from standardized primitives.

\textbf{Support for Complex Workflows:} Agent Spec supports advanced workflows and modular design patterns, facilitating the seamless development of sophisticated AI agents.

\textbf{Portable Execution Graphs:} Agent Spec enables execution graphs to be described in a way that is independent of the underlying runtime implementation, ensuring seamless portability across frameworks. Converters between the representations of other frameworks to Agent Spec simplifies the porting of previously developed agents to Agent Spec-compatible frameworks. A recent example is deprecation of AutoGen framework and corresponding migration to MS Agent Framework~\cite{autogenmigration}.

\subsubsection{Interoperability \& Compatibility}
\label{ref-023}


\textbf{SDK Support:} Agent Spec provides SDKs in various programming languages (starting with Python), supporting serialization and deserialization of agents into Agent Spec configurations. This simplifies development, deployment, and debugging across different environments.

\textbf{Robustness \& Consistency:} Agents defined in Agent Spec can be built on top of components that have been defined and validated within the Agent Spec representation, leading to higher reliability and predictability.

\textbf{Define Once, Run on Multiple Frameworks:} Agentic systems leveraging Agent Spec can run on various frameworks through runtime adapters, fostering open interoperability. This open specification allows developers to implement these adapters to support any framework, enabling agents to run in multiple environments.

\subsubsection{Unification across Frameworks}
\label{ref-024}


\textbf{Common Format:} Agent Spec's common interchange format simplifies integration of agents and enables easier agent sharing across platforms.

\textbf{Knowledge Exchange:} The unified declarative format that Agent Spec provides helps developers as well as maintainers of agents to efficiently exchange agent ideas and implementations.

\textbf{Conformance Test Suite:} The Agent Spec conformance test suite offers a comprehensive set of tests that cover the various functionalities and components available within the specification, such as Agent, Flow, BranchingNode, MapNode, and ToolNode. It includes illustrative validation examples of Agent Spec configurations and verifies that these configurations produce consistent execution outcomes and behaviors across multiple runtime environments. This approach ensures interoperability and correctness of implementations supporting the Agent Spec specification.

\section{Design Principles and Key Components}
\label{ref-005}

\begin{figure}[t]
\centering
\includegraphics[width=1\linewidth]{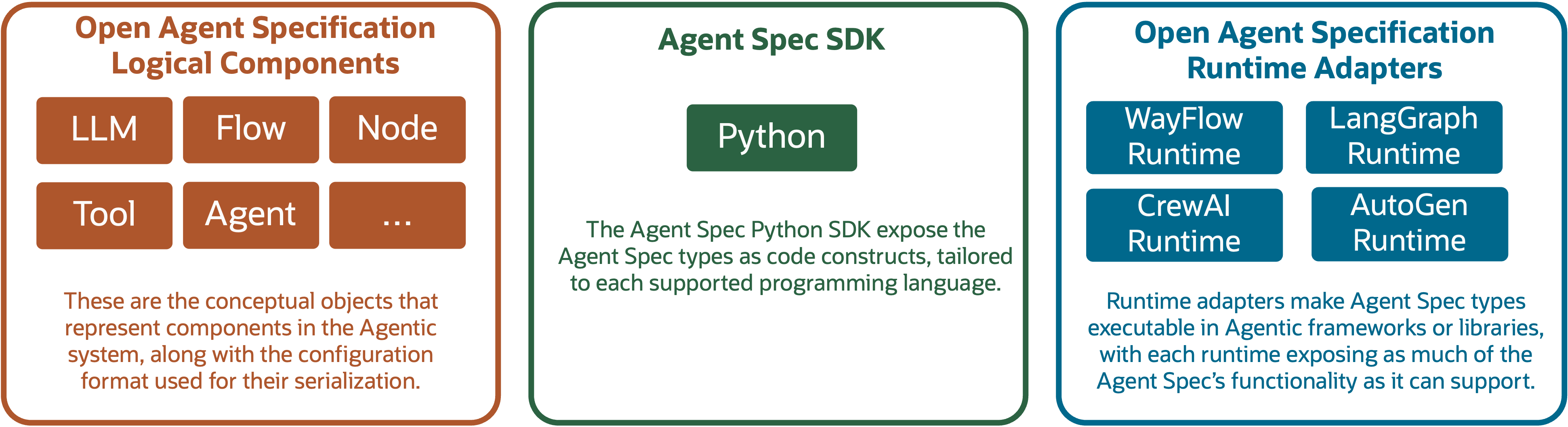}
\caption{Agent Spec's Design.}
\label{fig:AgentSpec_design}
\end{figure}

This section provides an overview of the declarative model defined by Agent Spec. It explains how Agent Spec operates across multiple frameworks and outlines design details including SDK structure, supported programming languages, agent definitions using the SDK, and serialization and deserialization processes for agents and workflows. Figure~\ref{fig:AgentSpec_design} provides an overview of the Agent Spec ecosystem.

Agent Spec is a portable, shareable, platform-agnostic configuration language that describes agents and agentic systems with sufficient fidelity for execution across runtimes. It defines the conceptual objects, called \textbf{components}, that make up agents in typical agentic systems, together with the properties that configure their behavior and semantics.

Runtimes implement these Agent Spec components for execution within their own frameworks. Agent Spec is supported by SDKs in multiple languages (starting with Python), enabling serialization and deserialization of Agents to and from JSON (or YAML), or direct creation from object representations while maintaining full specification compliance. As shown in \textbf{Table \ref{tab:AgentSpec-table}}, Agent Spec provides for AI agents what ONNX provides for ML models.

\subsection{Overview of Agent Specification}
\label{ref-006}
\begin{table}[ht]
\centering
\scriptsize 
\caption{Agent Spec serves a similar unifying role for AI agents as ONNX does for ML models.}
\begin{tabular}{p{1cm}p{3.0cm}p{3.0cm}}
\hline
\textbf{Feature} & \textbf{ONNX (ML Models)} & \textbf{Agent Spec (AI Agents)} \\ \hline
Scope & Unified representation of ML models, allowing portability across different deep learning frameworks. & Unified representation of AI agents and workflows, enabling them to run across diverse agent frameworks. \\ 
\hline
Portability & ONNX allows ML models to be trained in one framework (e.g., PyTorch, TensorFlow) and executed in another. & Agent Spec enables AI agents to be built in one framework (e.g., LangGraph \cite{LangGraph}, AutoGen \cite{autogen}, crewAI \cite{crewAI}) and run in another one. \\ 
\hline
Unification & Provides a common declarative format which is versioned as an Opset. & Provides a common agent configuration format with versioning support. \\ \hline
Extensibility & Supports various ML operations and optimizations. & Defines modular components for scalable agent systems. \\ \hline
\end{tabular}
\label{tab:AgentSpec-table}
\end{table}

The Open Agent Specification defines structure and behavior of the conceptual building blocks, called components, that make up agents in typical agent-based systems. The following sections describe each component in greater detail; the full specification can be found in \cite{language_spec_25_4_1}.

\subsection{Components}
\label{ref-007}

In Agent Spec, components are the building blocks which cover the commonly used elements in agentic systems, such as (but not limited to) Agents, LLMs, tools, etc.

Describing these components with specific types enables:
\begin{itemize}
\item \textbf{Type safety and GUI guidance:} For example, if a component property expects an LlmConfig, only compatible extensions can be connected, or valid connections can be highlighted to improve the agent-building experience in graphical interfaces.
\item \textbf{Static analysis and validation:} For instance, if a component is a \textit{Flow}, it must include a start node; if it is an \textit{Agent}, it must contain an LLM.

\item \textbf{Ease of programmatic use:} \texttt{Component} families have corresponding class definitions in the Agent Spec SDKs (see Section \ref{ref-025}). This allows consumers (e.g., execution environments or GUI editors) to work with concrete classes directly, rather than inferring which properties belong to which types.
\end{itemize}

\subsubsection{Base Component}
\label{ref-008}

The basic block in Agent Spec is a \texttt{Component}, which by itself can be used to describe any instance of any type, guaranteeing flexibility to adapt to future changes. It is important to note that Agent Spec does not need to encapsulate the implementation code it describes; it simply needs to be able to express enough information to instantiate a uniquely identifiable object of a specific type with a specific set of property values.

A \texttt{metadata} field contains all additional information that could be useful in different usages of the Agent Specification. For example, GUIs will require including some information about the components (e.g., position coordinates, size, color, notes), so that they will be able to visualize them properly even after an export-import process. Metadata is optional and set to \texttt{null} by default.

\paragraph{\textit{Symbolic references (configuration components)}}

When a component references another component entity (for example as a property value), there is a simple symbolic syntax to accomplish this, "\texttt{\$component\_ref:}" followed by the id of the other component.
\begin{lstlisting}[language=Python]
"$component_ref:{COMPONENT_ID}" 
\end{lstlisting}

This type of relationship is applicable to both Agent components as well as Flow components (a workflow component may also use an LLM or memory component).

We do not need a separate object for this type of reference, as the reference is explicitly assigned to some property or parameter.

\paragraph{\textit{Input/output schemas}}

Components might require input data to perform their task and produce output as result. These inputs and outputs must be declared and described in Agent Spec, so that users, developers and other components of the agent are aware of what is exposed as inputs and outputs respectively. Note that input/output schemas are not added directly to the base \texttt{Component} class, as there are a few cases where they do not apply (for example \texttt{LLMConfig} and edges; see the following sections for details).

\paragraph{\textit{Input and output properties}}
Input and output schemas define the "values" that different components can accept as input or produce as output. These values are referred to as "properties" based on \href{https://json-schema.org/}{JSON schema} standard~\cite{jsonschema}.

\paragraph{\textit{Inputs and outputs of nested components}}

\begin{figure}[H]
\caption{Illustration of input/output exposure in nested components.}
\includegraphics[width=1\linewidth]{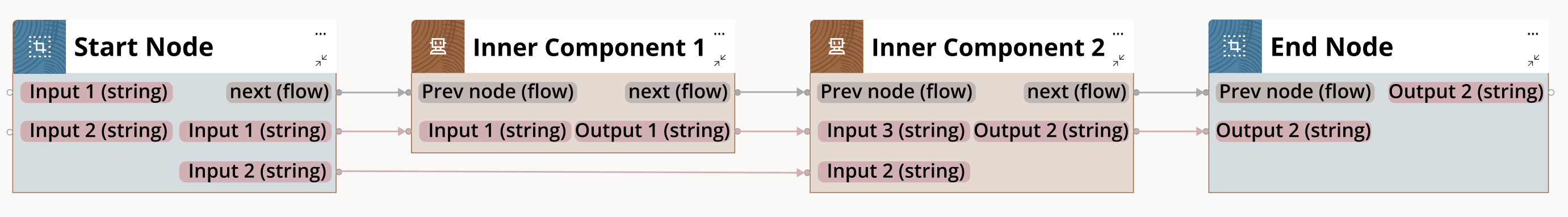}
\label{fig:2}
\end{figure}

In the case of nested components (for example, using an Agent component inside a Flow, or a Flow inside another Flow, or just a step in a Flow), the wrapping component is supposed to expose a (sub)set of the inputs/outputs provided by the inner components together with additional inputs/outputs it might generate as shown in Fig. \ref{fig:2}.

\vspace{-20pt} 

Each value listed in the inputs and outputs must replicate its corresponding JSON schema in every component that exposes it. Consequently, if a wrapper component makes available certain inputs or outputs from its internal components, these must also be explicitly included in the wrapper’s own input/output lists. This makes the specification of the components more verbose, but transparent and more readable. Parsers of the specification are required to ensure the consistency of the input/output schemas defined in it.

In the example above, the wrapping Flow component (see sections below for more details on Flows) requires two inputs that match those exposed by its nested components and produces one output derived from the inner components, along with an additional output computed internally.

\paragraph{\textit{Validation of specified inputs/outputs}}

Some input and output schemas are automatically generated by components based on their configuration. However, for the sake of clarity and readability of Agent Spec, all input and output descriptions must be explicitly included in the Agent Specification. Consequently, when a serialized specification is loaded by a runtime or SDK, the inputs and outputs configuration of each component must be validated against those automatically generated from its configuration (if applicable). 

\paragraph{\textit{Specifying inputs through placeholders}}

Some components can infer inputs from specific parts of their configurations. For example, certain nodes (e.g., the \texttt{LLMNode}) extract inputs from their configuration attributes, allowing their values to be dynamically adapted based on previous computations. Users can define placeholders within these attributes by enclosing the name of a property in double curly brackets.
The node automatically infers the inputs by interpreting those configurations and the actual value of the input/output will be substituted at execution time.
Whether an attribute accepts placeholders depends on the definition of the component itself.
For example, setting the \texttt{prompt\_template} configuration of an \texttt{LLMNode} to "You are a great assistant. Please help the user with this request: \{\{request\}\}" will generate an input called "request" of type string.

\subsubsection{Agent}
\label{ref-009}

The \texttt{Agent} is the top-level component that holds shared resources such as conversation memory and tools. It also serves as the entry point for interactions with the agentic system. Its primary goal is to populate the values of all properties defined in its output attribute, either through user interaction or autonomous computation.

Defining Agents as standalone components is essential, as it enables the same Agent component to be reused across multiple Flows or within more complex agent architectures, without the need to duplicate its definition.

\subsubsection{LLM}
\label{ref-010}

Large Language Models (LLMs) are used within agents as generative components. To integrate them effectively, they must be flexibly configurable through parameters such as connection details, model identifiers, generation settings, and other options.

By default, only the generation parameters required for prompting the LLM need to be specified. Specific extensions of \texttt{LLMConfig} can be derived to support the most common model providers. This approach is not limited to locally hosted models; it also supports the configuration of external API endpoints, such as those offered by OpenAI, Anthropic, Mistral, and others.

\subsubsection{Tool}
\label{ref-011}

A tool is a procedural function or a Flow that can be made available to an Agent component to execute and perform tasks. The Agent can decide which tool to invoke based on the tool's signature and description. In a Flow context, a node must be explicitly configured to call a specific tool.

A novelty in Agent Spec is the classification of tools according to where their actual functionality is executed. Four corresponding types are defined:

\begin{itemize}[noitemsep, topsep=0pt]
\item \textbf{ServerTools} are executed within the same runtime environment as the agent itself. The definitions of these tools therefore must be available in the agent's environment. Such tools are typically limited in number and functionality.
\item \textbf{ClientTools} are not executed by the runtime. Instead, they run in the client environment, which is responsible for executing the tool and returning the results to the runtime (similar to OpenAI’s function calling model).
\item \textbf{RemoteTools} are executed in an external environment and triggered through an RPC or REST call initiated by the agent runtime.
\item \textbf{MCPTools} enable execution of tools provided by a MCP server in accordance with the MCP specification \cite{mcp}.
\end{itemize}

Agent Spec does not prescribe how these tool types should be implemented. Rather, it provides a representation format that allows them to be correctly interpreted and exchanged across different platforms and programming languages.

Conceptually, a \texttt{Tool} is a function that accepts parameters (inputs), performs a transformation, and returns a result (outputs). Each tool has a name and a description, which can provide context to an LLM when deciding which tool to use for a given task. Consequently, a \texttt{Tool} is defined as an extension of \texttt{ComponentWithIO}.

Tools may produce multiple outputs. In such cases, the expected return value is a dictionary, where each key corresponds to the name of an output property defined in the tool's schema. The runtime should parse the dictionary to extract the different outputs and bind them correctly.

While \texttt{ServerTool} and \texttt{ClientTool} do not require additional parameters beyond their standard definition, a \\\texttt{RemoteTool} must include the necessary details for invoking the remote call.

Finally, an important security consideration in Agent Spec is the exclusion of arbitrary code from the specification. The Agent Specification only provides a structured description of each tool, including its attributes, inputs, outputs, and metadata but does not contain any executable code.

\subsubsection{Flow}
\label{ref-012}

Flows (or graphs) are directed, potentially cyclic workflows. They can be regarded as "subroutines" that encapsulate structured sequences of operations. As such, Flows offer greater determinism and reliability compared to purely agent-driven components at the cost of reduced flexibility.

Each Flow may require zero or more inputs (corresponding to the inputs expressed on the starting node of the graph) and may produce zero or more outputs (represented by the terminal node of the graph - note that a graph might contain multiple terminal nodes in cases involving branching logic).

\texttt{Flow} objects specify the directed graph's entry point, its nodes, and the edges connecting them. The individual components that compose a Flow are described in detail in the following sections.

A Flow exposes all inputs defined in its \texttt{StartNode} and all outputs that are available across its \texttt{EndNode} objects.

\paragraph{Conversation}

At the core of a conversational agent’s execution lies the conversation itself. The conversation is implicitly propagated across all components within the Flow and maintains the collection of messages generated throughout execution. Each node in a Flow, as well as each Agent, might append messages to this conversation as needed. Every message contains, at minimum, the following attributes: role (e.g., system, agent, user), content, sender, and recipient.

\paragraph{Input/Output Data Space}

Alongside the conversation, flows have another data space composed of inputs and outputs generated by nodes.
When a node generates an output, this is exposed by the node itself, so that other nodes can consume it as input, if needed.
The input-output connections, called \textit{data flow}, can be defined through explicit data relationships, or through implicit, name-based property matching, as depicted in the following sections.

\paragraph{Relationships / Edges}

Within Flows, Agent Spec distinguishes between two categories of relationships (or edges): those governing data flow and those governing control flow. This distinction enables Agent Spec to unambiguously represent complex execution patterns such as conditional branches and loops, which are difficult to model using data flow alone.

Relationships are defined exclusively within the context of Flows. Each relationship expresses a directed transition from a source node to a target node.

\subsubsection{ControlFlowEdge}
\label{ref-013}

A control flow relationship specifies a potential transition from one branch of a node to another. The actual transition followed during execution is determined by the implementation of the corresponding node.

The attribute \texttt{from\_branch} of a \texttt{ControlFlowEdge} is set to \texttt{null} by default, indicating that the edge connects to the node's default branch (i.e., \textit{next}).

\subsubsection{DataFlowEdge}
\label{ref-014}

In an input/output system, a component identifier alone is insufficient to fully specify a data relationship. Therefore, Agent Spec defines explicit data flow relationships to describe how the output property of one node is mapped to the input property of another.

A component might accept multiple input parameters and/or produce multiple output values. Hence, it is necessary to define which output or reference is associated with each input, and conversely, how each output is consumed.

Destinations (the inputs of the target node) can be connected to outputs from another node or assigned to static values. Conversely, sources (the outputs of the source node) may be connected to inputs of subsequent nodes or left unconnected if unused.

\paragraph{Connecting multiple data edges}

\begin{figure}[t]
\centering
\begin{subfigure}[b]{\linewidth}
\centering
\includegraphics[width=\linewidth]{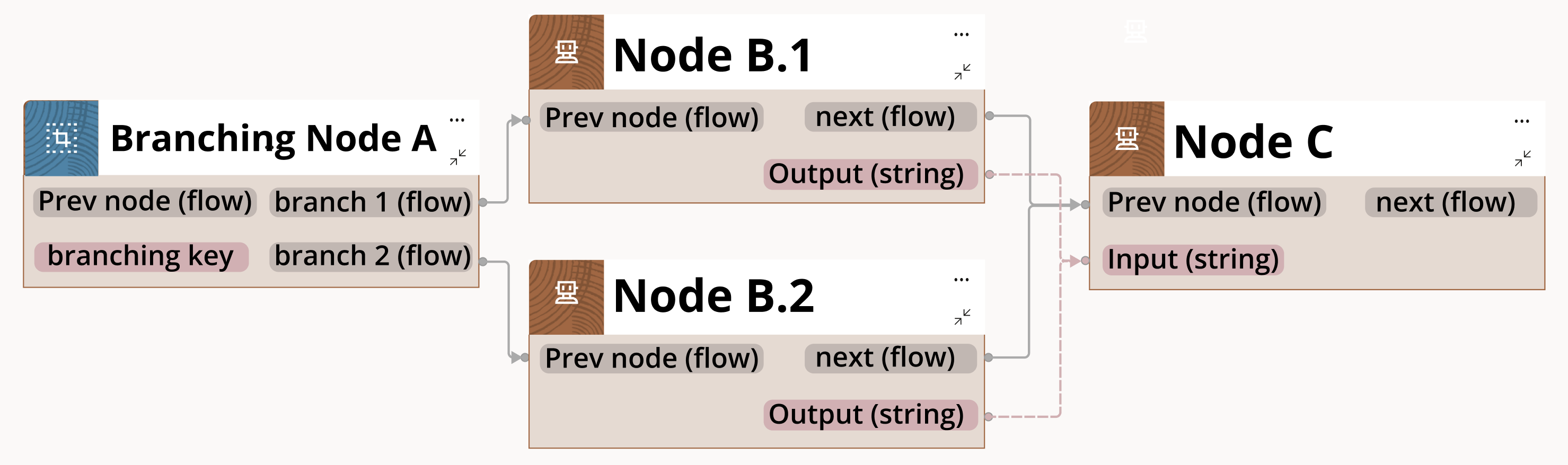}
\caption{Control flow with two branches that converge into the same node with a property filled based on the branch taken.}
\label{fig:Ng1}
\end{subfigure}
\vspace{0.6em}
\begin{subfigure}[b]{\linewidth}
\centering
\includegraphics[width=\linewidth]{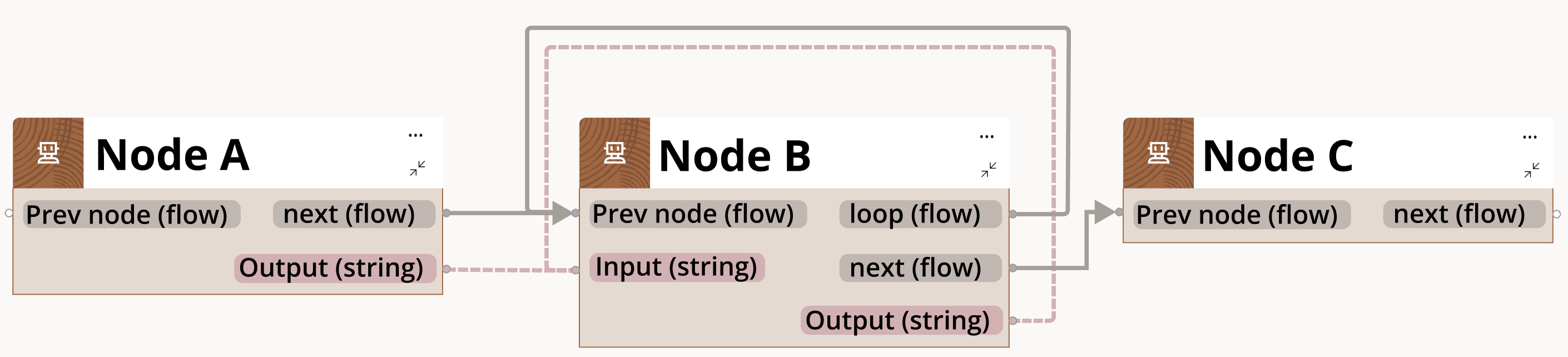}
\caption{Control flow with a self-loop and a property updated by consequent execution of the same node}
\label{fig:Ng2}
\end{subfigure}
\caption{Control- and data-flow relationships in Agent Spec. Solid lines denote control-flow transitions, while dotted lines indicate data propagation.}
\label{fig:3}
\end{figure}

In the definition of a Flow, it is permissible to connect two or more distinct data outputs to the same input as shown in Fig. \ref{fig:3}.
In such cases, the value produced by the most recently executed node among those connected to the input takes precedence. Conceptually, this behavior is analogous to exposing the input as a public variable, where each connected node can update its value upon execution.

In contrast, control flow edges define the set of permitted next-step transitions. Multiple outgoing connections from the same control flow branch are not permissible, ensuring unambiguous execution order. It should also be noted that parallelism is not enabled through control flow edges.

\paragraph{Optionality of data flow relationships}
\label{ref-optionality-data-relationships}

Some frameworks do not require the explicit specification of data flow, as they assume that all properties are publicly exposed within a shared "variable space" accessible to any component. In such systems, read and write operations are determined purely by name-based resolution.

In practice, this means that:

\begin{itemize}
\item When a component defines an input with a given name, it retrieves the value of the variable with the same name from the shared variable space.
\item When a component defines an output with a given name, it writes (or overwrites, if already present) the corresponding variable in the shared variable space.
\end{itemize}

Users can adopt this type of I/O mechanism in Agent Spec by setting the \texttt{data\_flow\_connections} parameter of a \texttt{Flow} to \texttt{None}. In this configuration, the system follows the name-based approach described above, using the shared variable space as the public medium for publishing and accessing values.

It is worth noting that this name-based strategy can be expressed equivalently by explicitly defining the corresponding data flow connections by connecting all outputs and inputs that share the same name and following the control flow to account for overwrites. However, the converse is not true: an explicitly defined data flow configuration cannot always be reduced to a purely name-based representation.

\subsubsection{Node}
\label{ref-015}

A Node represents a vertex within a Flow and is derived from the \texttt{ComponentWithIO} class. Agent Spec provides a standard library of node types designed to facilitate workflow orchestration, enabling efficient development and seamless integration across a wide range of applications.

The main node types include:
\begin{itemize}
\item \texttt{LLMNode}: Utilizes a large language model (LLM) to generate text from a given prompt.
\item \texttt{APINode}: Performs an API call based on the specified configuration and inputs.
\item \texttt{AgentNode}: Executes a multi-turn interaction with an \texttt{Agent} component, supporting modular agent design and reuse.
\item \texttt{FlowNode}: Executes a \texttt{Flow}, enabling hierarchical structuring and reuse of Flows within more complex architectures.
\item \texttt{MapNode}: Performs a map-reduce operation over an input collection, applying a specified \texttt{Flow} to each element in the collection.
\item \texttt{StartNode}: Defines the entry point of a \texttt{Flow}.
\item \texttt{EndNode}: Defines the termination point of a \texttt{Flow}.
\item \texttt{BranchingNode}: Enables conditional branching based on the value of a specified input.
\item \texttt{ToolNode}: Invokes and executes a defined \texttt{Tool}.
\item \texttt{InputMessageNode}: Interrupts temporarily the execution to retrieve user input.
\item \texttt{OutputMessageNode}: Appends an agent message to the conversation.
\end{itemize}
A comprehensive description of each node type is provided in the Agent Spec documentation, available at \cite{language_spec_25_4_1}. 

\subsection{Open Agent Specification Model}
\label{ref-016}

The primary objective of Agent Spec is to remain agnostic with respect to any agentic framework and the programming language used to build and execute an agent. Consequently, an agent represented in Agent Spec should be exportable to, and importable from a common, framework-independent serialization format.
To achieve this, Agent Spec adopts \href{https://www.json.org/json-en.html}{JSON} as the designated language for the serialization of its components.

All building blocks within Agent Spec are explicitly designed to be trivially serializable. Thus, the Agent Specification of any component can be directly obtained by serializing its definition according to the JSON standard. Each serialized component includes its set of attributes, along with an additional field, \texttt{component\_type}, which is used during deserialization to infer the appropriate component type.

As defined by the Open Agent Specification, each component may be declared only once within a representation and subsequently referenced symbolically, following the rules outlined in the Symbolic References section of this article. Illustrative examples of component serializations in JSON format are available at \cite{language_spec_25_4_1}.

\subsection{How Agent Spec Works Across Frameworks}
\label{ref-017}


The primary objective of Agent Spec is to define behavioral patterns for the various components that can be implemented according to the specific characteristics of different agentic frameworks. A runtime environment developed for a particular framework should be capable of reading the serialized configuration of a component (such as a Flow or an Agent) and executing its logic in accordance with the behaviors prescribed by Agent Spec. Every Agent Spec configuration includes a top-level \texttt{agentspec\_version} property specifying the specification version it follows.

Alternatively, if the agentic framework used to implement the runtime provides its own serialization mechanism, a translation layer could be introduced. This layer can convert an Agent Spec configuration into the equivalent framework-specific representation, allowing the runtime to execute the agent directly using the translated format.

\subsection{Implementation Details}
\label{ref-018}
To facilitate the programmatic construction of framework-agnostic agents, Agent Spec supports the development of dedicated Software Development Kits (SDKs). Our reference SDK, \texttt{PyAgentSpec}, also includes a plugin architecture that allows teams adopting Agent Spec to define and integrate new component types within their configurations as needed. A runtime adapter implementation makes Agent Spec component abstractions executable by providing the behavioral logic defined in the Agent Spec language. Beyond simple translation, these runtime adapters act as “compiler” for Agent Spec, converting its declarative specifications into executables for target frameworks and bridges high-level declarative design with low-level execution semantics. Please see Appendix \textbf{A} for an example of an agentic flow in PyAgentSpec and Appendix \textbf{B} for details on the reference Agent Spec SDK, \texttt{PyAgentSpec}, the plugin architecture and runtime adapters.

\section{Evaluation \& Benchmarking}
\label{ref-025}

Agent Spec supports key aspects of the productization of agent-driven solutions, including the comparison and evaluation of agent task performance across different execution frameworks and model configurations, as well as optimization of execution costs and efficiency across different environments.

By encompassing widely accepted concepts in agentic systems, such as workflows and agents, Agent Spec provides a unifying abstraction that enables developers to map these concepts consistently across diverse execution frameworks. This facilitates direct performance benchmarking and comparative analysis of framework-specific task execution. Furthermore, compute optimization is also possible, as Agent Spec enables hardware-optimized and framework-optimized implementation of common agentic design patterns. Please see Appendix \textbf{C} for additional details on real-world use cases \& business applications. 

We note that the goal of this evaluation is not to outperform existing implementations of agents, but to show that Agent Spec enables cross-framework portability and reusability of non-trivial, competitive agents.

\subsection{Evaluation methodology}
We evaluate Agent Spec on three popular benchmarks: \textbf{SimpleQA Verified}~\cite{haas2025simpleqaverified}, \textbf{$\tau^2$-Bench}~\cite{barres2025tau2} and \textbf{BIRD-SQL}~\cite{li2024can}. For each benchmark, we select one or more Agent Spec JSON configurations and execute the same specification across four framework runtimes: AutoGen, CrewAI, LangGraph, and WayFlow. This approach allows direct comparison of behavioral consistency, correctness, and latency across heterogeneous frameworks. In this evaluation, the same agent utilizing identical tools is instantiated in AutoGen, CrewAI, LangGraph, and WayFlow. 

We report \texttt{F1} scores for \texttt{SimpleQA Verified} (exact-match over verified answers), \texttt{EX\%} for \texttt{BIRD-SQL} (execution accuracy for generated SQL compared to ground-truth SQL), and \texttt{Pass\^{}k} for \texttt{$\tau^2$-Bench} (probability to get all k attempts correct out of k). Query time is reported as mean ± standard deviation over all records.

\paragraph{SimpleQA Verified.} We evaluate on the full set of 1,000 questions using two representative assistants: (i) a ReAct-style agent equipped with a web-browsing tool, and (ii) a flow-based agentic RAG solution~\footnote{Details about the flow is provided in Appendix \textbf{D.1}.} that applies question decomposition, web-browsing, and self-reflection to answer complex user queries\footnote{Note that the browse tool makes the task~\cite{haas2025simpleqaverified} easier, but it is acceptable since we are demonstrating Agent Spec’s functional effectiveness instead of competing on the benchmark leaderboard.}. We also include the performance of a single LLM call as a baseline. 

\paragraph{Bird-SQL} We evaluate execution accuracy on the 1,534 questions from BIRD-SQL dev with two types of assistants. The first assistant is a ReAct-style agent equipped with a tool to validate that a query can be executed, and a "thinking" tool to give opportunities to the agent to think between tool calls. The second assistant is a flow-based assistant~\footnote{Details about the flow is provided in Appendix \textbf{D.2}.} composed of a planning step, a step to generate the SQL query and a final post-processing step to self-reflect on the generated query. We also include the performance of a single LLM call as a baseline. 

\paragraph{$\tau^2$-Bench.} A simulation environment designed to evaluate conversational customer service agents in a dual-control setting, where both the agent and a user simulator interact within realistic domains such as airline, retail, and telecom. For each domain the benchmark provides a set of tools to configure the agent, and requests are given to the agent via a user simulation. Both the actual actions of the agent and the information communicated to the user are scored by the benchmark. We assess performance in the retail domain using a max-steps limit of 40 and \texttt{GPT-4.1} as the user model. For the ReAct-style agent, we evaluate both \texttt{GPT-4.1} and \texttt{GPT-4.1 mini} as the underlying models.

\subsection{Results and Insights}
Agent Spec successfully instantiates functional agents across all tested frameworks, confirming consistent behavior and validating its framework-agnostic portability \& reusability. This confirms that a single declarative specification can reproduce agent execution across heterogeneous runtimes. 

\paragraph{SimpleQA Verified.} On \texttt{SimpleQA} (Table~\ref{tab:simpleqa-bench-table}), the large gap between simple prompting of GPT-4.1 mini and the ReAct agent implementation (both based on the same model) shows that all frameworks gain substantially from the use of web search tools. Even a compact model like GPT-4.1 mini can successfully orchestrate ReAct agents. Using a more powerful model to orchestrate the agent does not yield meaningful quality improvements. Latency differences mainly arise because the Llama model runs locally while GPT-4.1 mini depends on a remote API with variable response times. The flow-based solution leads to higher accuracy thanks to decomposition and self-reflection. High variance in time per query is connected to the different number of reasoning iterations (see Appendix \textbf{D.1} for more information on the flow implementation).

Differences in ReAct agent performance reveal interesting insights into each framework's implementation, which in turn affect the final outcomes.
\begin{itemize}
\item WayFlow and LangGraph show similar behaviors, with WayFlow being slightly more accurate but slower than LangGraph.
\item AutoGen, a multi-agent system–oriented framework, performs less effectively in this single-agent scenario. Its lack of support for non-native tool calling further limits its performance compared with other frameworks using the Llama model.
\item CrewAI achieves the highest F1 score but at the cost of significantly higher response times between 55\% (with \texttt{GPT-4.1 mini}) to 140\% slower (with \texttt{Llama 3.3 70B}) than competitors and higher setup complexity, as it requires additional directives to create the agent’s task.
\end{itemize}

\begin{table}[ht]
\centering
\setlength{\tabcolsep}{1.5pt} 
\renewcommand{\arraystretch}{1.1} 
\caption{Benchmark results on SimpleQA Verified. Flow-based results for AutoGen and CrewAI are not reported since they do not support non-agentic flow execution.}
\label{tab:simpleqa-bench-table}
\begin{tabular}{@{}cccc@{}}
\toprule
\textbf{Agent Setup}                & \textbf{Runtime}    & \textbf{F1-score (\%)}        & \textbf{Time/query (s)}   \\
\midrule
GPT-4.1 mini                  & -                   & 17.8              & $1.06 \pm 0.41$      \\
\midrule
\multirow{4}{*}{\shortstack{ReAct Agent\\(GPT-4.1 mini)}} & WayFlow   & 79.0    & $6.22 \pm 5.24$      \\
                              & LangGraph           & 75.5              & $4.57 \pm 6.21$      \\
                              & AutoGen             & 67.9              & $4.64 \pm 1.64$      \\
                              & CrewAI              & \textbf{81.1}     & $7.21 \pm 6.74$      \\
\midrule
\multirow{4}{*}{\shortstack{ReAct Agent\\(Llama 3.3 70B)}} & WayFlow  & 79.2    & $4.60 \pm 0.92$      \\
                              & LangGraph           & 77.6              & $4.10 \pm 0.91$      \\
                              & AutoGen             & 54.9              & $3.38 \pm 0.31$      \\
                              & CrewAI              & \textbf{82.3}     & $8.12 \pm 3.84$      \\
\midrule
\multirow{2}{*}{\shortstack{Flow-based\\agentic RAG\\(GPT-4.1 mini)}} & WayFlow   & \textbf{88.1}     & $23.19 \pm 43.25$    \\
                              & LangGraph           & 85.3              & $23.21 \pm 55.53$  \\[1ex]
\bottomrule
\end{tabular}
\end{table}

\paragraph{BIRD-SQL.} On \texttt{BIRD-SQL} (Table~\ref{tab:bird-bench-table}), the performance gap between the different frameworks is more narrow compared to \texttt{SimpleQA} and \texttt{$\tau^2$-Bench}, which is to be expected given that NL2SQL tasks require short trajectories to be solved. CrewAI is still significantly slower compared to the three other runtimes. Giving tools to the Agent does not improve over the baseline performance but the Plan-Generate-Reflect flow consistently outperforms the baseline.

\begin{table}[ht]
\centering
\caption{Benchmark results on BIRD-SQL. Flow-based results for AutoGen and CrewAI are not reported since they do not support non-agentic flow execution.}
\label{tab:bird-bench-table}
\begin{tabular}{@{}cccc@{}}
\toprule
\textbf{Agent Setup}                & \textbf{Runtime}    & \textbf{EX (\%)}        & \textbf{Time/query (s)}   \\
\midrule
GPT-4.1 mini                  & -                   & 50.1              & $4.26 \pm 2.25$      \\
\midrule
GPT-4.1                       & -                   & 54.4              & $2.93 \pm 4.68$      \\
\midrule
\multirow{4}{*}{\shortstack{ReAct Agent\\(GPT-4.1)}}     & WayFlow   & 54.1    & $7.16 \pm 2.72$      \\
                              & LangGraph           & 53.7              & $5.68 \pm 3.64$      \\
                              & AutoGen             & 54.1              & $6.06 \pm 2.97$      \\
                              & CrewAI              & \textbf{54.3}     & $12.20 \pm 6.62$      \\
\midrule
\multirow{2}{*}{\shortstack{NL2SQL Flow\\(GPT-4.1)}} & WayFlow   & 55.5     & $9.45 \pm 5.56$    \\
                              & LangGraph           & \textbf{55.9}       & $7.54 \pm 4.11$  \\

\bottomrule
\end{tabular}
\end{table}

\paragraph{$\tau^2$-Bench.} 
Performance differences on \texttt{$\tau^2$-Bench} (Table~\ref{tab:tau2-bench-table}) stem from variations in prompt-handling strategies across frameworks which wrap the system prompt and add additional instructions in distinct ways. These frameworks also differ in how they request tool usage from the hosted model. Such differences are influenced by how each model is hosted and how it manages API calls and responses.

Notably, Agent Spec exposes these variations, enabling consistent cross-framework evaluation. Additionally, we observe the lowest scores for the CrewAI runtime, which can be attributed to its design focus on multi-agent coordination for single-task problem-solving, rather than on conversational interactions with users as evaluated by $\tau^2$-Bench.

\begin{table}[ht]
\centering
\setlength{\tabcolsep}{1.5pt} 
\renewcommand{\arraystretch}{1.1} 
\caption{Benchmark results on $\tau^2$-Bench. Experiments are grouped by the base LLM used.}
\label{tab:tau2-bench-table}
\begin{tabular}{@{}cccc@{}}
\toprule
\textbf{Agent Setup} & \textbf{Runtime} & \textbf{Pass\^{}1 / \^{}4 (\%)} & \textbf{Time/query (s)} \\
\midrule
\multirow{4}{*}{\shortstack{ReAct Agent\\(GPT-4.1)}} &
  WayFlow      & 64.7 / 43.0           & $43.0 \pm 13.8$ \\
& LangGraph    & 71.5 / 51.8           & $37.5 \pm 12.4$ \\
& AutoGen      & \textbf{73.5} / \textbf{52.6} & $55.5 \pm 17.8$ \\
& CrewAI       & 62.5 / 38.6           & $62.7 \pm 25.4$ \\
\midrule
\multirow{4}{*}{\shortstack{ReAct Agent\\(GPT-4.1 mini)}} &
  WayFlow      & 60.1 / 32.5           & $40.9 \pm 12.7$ \\
& LangGraph    & \textbf{60.7} / 32.5  & $38.4 \pm 17.4$ \\
& AutoGen      & 58.3 / \textbf{36.0}  & $47.4 \pm 15.0$ \\
& CrewAI       & 51.5 / 22.8           & $56.1 \pm 25.5$ \\
\bottomrule
\end{tabular}
\end{table}


\section{Conclusion \& Potential Enhancements}
\label{ref-029}

This paper introduced \textbf{Agent Spec}, a declarative, framework-agnostic specification for defining AI agents and workflows. By decoupling design from execution, it enables define-once, run-anywhere portability and interoperability across agentic frameworks. We formalized its core components and semantics, and released toolsets: a Python SDK (PyAgentSpec), a reference runtime (WayFlow), and adapters for frameworks (e.g., LangGraph and AutoGen).

We empirically show that Agent Spec serves as an \textbf{Evaluation harness}, enabling consistent comparison of agent and flow patterns across frameworks. We outline potential extensions to Agent Spec’s reusability, interoperability, and developer experience in Appendix \textbf{E}. Planned enhancements include compiler-style optimizations such as fusing consecutive \texttt{LLMNode}s and using structured generation to reduce redundant calls. Looking ahead, we envision Agent Spec as a foundation for standardizing agent design \& evaluation. 

\bibliography{bibliography}
\bibliographystyle{unsrt}

\newpage
\appendix

\section{Simple Agentic Flow in PyAgentSpec}
\addcontentsline{toc}{section}{Appendix}
\label{sec:agentflow_example}
The following example illustrates the definition of a simple agentic Flow in PyAgentSpec that prompts an LLM with a user-provided input and returns the model's generated output.

\begin{lstlisting}[caption={Simple Agentic Flow in PyAgentSpec}, language=Python]
from pyagentspec.property import Property 
from pyagentspec.flows.flow import Flow 
from pyagentspec.flows.edges import ControlFlowEdge, DataFlowEdge 
from pyagentspec.flows.nodes import LlmNode, StartNode, EndNode 
from pyagentspec.llms import VllmConfig 
llm_config = VllmConfig( 
    name="<Your Model Name Here>",
    url="<url.of.your.llm.deployment:port>", 
    model_id="<provider/model-identifier>", 
) 
prompt_property = Property( 
    json_schema={"title": "prompt", "type": "string"} 
) 
llm_output_property = Property( 
    json_schema={"title": "llm_output", "type": "string"} 
) 
start_node = StartNode(name="start", inputs=[prompt_property]) 
end_node = EndNode(name="end", outputs=[llm_output_property]) 
llm_node = LlmNode( 
    name="simple llm node", 
    llm_config=llm_config, 
    prompt_template="{{prompt}}", 
    inputs=[prompt_property], 
    outputs=[llm_output_property], 
) 
flow = Flow( 
    name="Simple prompting flow", 
    start_node=start_node,
    nodes=[start_node, llm_node, end_node], 
    control_flow_connections=[ 
        ControlFlowEdge(name="start_to_llm", from_node=start_node, to_node=llm_node), 
        ControlFlowEdge(name="llm_to_end", from_node=llm_node, to_node=end_node), 
    ], 
    data_flow_connections=[ 
        DataFlowEdge( 
            name="prompt_edge", 
            source_node=start_node, 
            source_output="prompt", 
            destination_node=llm_node, 
            destination_input="prompt", 
        ), 
        DataFlowEdge( 
            name="llm_output_edge", 
            source_node=llm_node, 
            source_output="llm_output", 
            destination_node=end_node, 
            destination_input="llm_output" 
        ), 
    ], 
)    
\end{lstlisting}

To export the Flow that was just created in JSON format, use the following code.

\begin{lstlisting}[caption={Export flow to JSON}, language=Python]
serializer = AgentSpecSerializer()
serialized_flow = serializer.to_json(flow)
\end{lstlisting}

\section{Appendix - Implementation Details}
\label{ref-appendix-b}
\subsection{Agent Spec SDKs}
\label{ref-019}
Agent Spec SDKs are expected to provide two core functionalities:

\begin{itemize}
\item Construction of Agent Spec component abstractions through the implementation of the corresponding interfaces, in full conformity with the Agent Spec standard.
\item Import and export of these abstractions to and from their serialized representations in JSON format.
\end{itemize}

As part of the Agent Spec initiative, we provide a reference implementation in Python, called PyAgentSpec~\cite{pyagentspec}.
PyAgentSpec enables developers to construct Agent Spec-compliant agents directly in Python. Users can define their own assistants by instantiating component classes that mirror the interfaces and behaviors prescribed by Agent Spec, and subsequently export these definitions to JSON.


\subsection{Agent Spec SDK Plugins}
\label{sec:plugins}

A plugin can extend the standard set of Agent Spec components by defining additional component types and specifying the corresponding serialization and deserialization logic required to handle them.

These plugins enable support for components not yet included in the current version of Agent Spec, such as those related to memory.

The abstract interface for implementing a serialization plugin is shown below; the deserialization interface follows the same design principles.

\begin{lstlisting}[caption={Interface for serializing an Agent Spec plugin}, language=Python]
from abc import ABC, abstractmethod
from typing import Any, Dict, List

from pyagentspec.component import Component
from pyagentspec.serialization.serializationcontext import SerializationContext

class ComponentSerializationPlugin(ABC):
    """Base class for Component serialization plugins."""

    @property
    @abstractmethod
    def plugin_name(self) -> str:
        """Return the plugin name."""
        pass

    @property
    @abstractmethod
    def plugin_version(self) -> str:
        """Return the plugin version."""
        pass

    @abstractmethod
    def supported_component_types(self) -> List[str]:
        """Indicate what component types the plugin supports."""
        pass

    @abstractmethod
    def serialize(
        self, component: Component, serialization_context: SerializationContext
    ) -> Dict[str, Any]:
        """Serialize a component that the plugin should support."""
        pass
\end{lstlisting}

\subsection{Agent Spec Runtime Adapters}
\label{ref-020}

There are multiple ways to implement runtimes for Agent Spec. Ideally, a runtime corresponds to an agentic framework that natively supports Agent Spec, i.e., it can directly read an Agent Spec JSON specification and instantiate an equivalent executable component.

As a reference example, we provide a runtime implementation called \texttt{WayFlow}\footnote{\url{https://github.com/oracle/wayflow}}, a powerful and intuitive Python library for constructing sophisticated AI-powered assistants.
WayFlow includes a library of prebuilt plan steps to streamline assistant creation, promotes component reusability, and supports rapid prototyping.
The framework implements both serialization and deserialization interfaces, enabling the import and export of Agent components that conform to Agent Spec.
It serves as the reference runtime for Agent Spec, offering native support for all Agent Spec Agents and Flows.

In cases where an agentic framework does not natively support Agent Spec, a translation layer that transforms the Agent Specifications into a framework-specific representation could be adopted.
This layer, which we call Agent Spec Runtime Adapter, loads an Agent Specification and converts it into an equivalent representation compatible with the target agentic framework, that can be finally executed.

We provide runtime adapters\footnote{\url{https://github.com/oracle/agent-spec/tree/main/adapters}} for some of the most popular agentic frameworks, including, but not limited to, \href{https://www.langchain.com/langgraph}{LangGraph} and Microsoft \href{https://microsoft.github.io/autogen/stable/index.html}{AutoGen}. These adapters leverage the Agent Spec Python SDK to load specifications and translate Agent Spec components into their respective framework-specific equivalents.

\section{Appendix - Use Cases \& Applications}
\label{ref-appendix-c}

A common use case for Agent Spec is the portability of agent-based assistants across different frameworks. Consider, for instance, a team that has implemented agent-based workflows using AutoGen. As business requirements and process needs evolve (or framework is deprecated~\cite{autogenmigration}), the team evaluates alternative frameworks. However, due to AutoGen's specific architecture, migrating these workflows to another agentic framework can be complex and typically entails substantial rewriting overhead.

In such cases, Agent Spec enables the development of agentic systems, tailored to specific business units, using a declarative representation. This representation can then be converted into framework-specific formats or executed flexibly across multiple frameworks via appropriate runtime adapters. This approach facilitates direct comparison of task performance and execution costs across frameworks. Consequently, employing Agent Spec throughout the agent design and deployment lifecycle significantly reduces complexity, implementation effort, maintenance cost, and potential inconsistencies.

Another key use case involves the development and deployment of Agents across heterogeneous compute environments, such as cloud platforms and on-premises data centers. In cloud-based systems, Agents might run in virtualized or serverless configurations as part of a scalable architecture, whereas in on-premises contexts, they might operate within traditional application servers. Agent Spec’s framework and runtime-agnostic design enables development of the Agent in a local development environment, followed by flexible deployment across either of these scenarios.

An illustrative example of this interoperability is available at \url{https://oracle.github.io/agent-spec/howtoguides/howto_execute_agentspec_across_frameworks.html} which demonstrates how an agent defined with Agent Spec can be executed across different frameworks.

\section{Appendix - Agentic Flows}
\label{ref-appendix-d}

\subsection{RAG Agentic Flow}
\label{ref-032}

The Flow-based Agentic RAG used for SimpleQA is an advanced question-answering system that adapts its reasoning strategy to the complexity of each query. The assistant begins with a basic Retrieval-Augmented Generation (RAG) pipeline for straightforward questions and routes to iterative, modular reasoning involving question decomposition, search, result deduplication, and answer synthesis for more complex or open-ended queries.

Initially, the user's query is embedded and sent to a web search tool, which retrieves a set of relevant results. These results are synthesized into a preliminary answer, which is judged by an LLM to assess if it is likely sufficient to answer the initial user question. If the system determines that this initial response adequately answers the user's question, it is returned immediately, minimizing latency.

If the response is missing, incomplete or lacks confidence, the system automatically transitions to a more sophisticated, iterative research mode. This pipeline begins by constructing a comprehensive context from the user's original question, any prior sub-questions, retrieved information, and feedback from earlier cycles. It then decomposes the complex query into focused sub-questions, which helps target web search and filter out irrelevant details. For each sub-question, the system retrieves new results, deduplicates overlapping content, and reassesses previously gathered evidence. The assistant then synthesizes a detailed answer and an evaluation component determines whether the answer is now comprehensive; if so, it finalizes and returns the response. Otherwise, the refinement loop continues until the assistant achieves a confident response or a maximum number of iterations is reached.

\subsection{NL2SQL Agentic Flow}
\label{ref-033}
This flow incorporates a structured sequence of agentic components to improve reasoning and reliability. It begins with an LLM planning node that interprets the user's query and formulates a high-level strategy. The plan is then executed by a ReAct-style agent equipped with external tools for retrieval, computation, or other actions. Finally, a post-processing LLM node reviews and refines the agent's output to verify correctness, check for specified errors, and format the final response before returning it to the user. This design was introduced because vanilla ReAct agents based on GPT-4.1 often fail to autonomously generate explicit plans or to perform final verification steps, leading to inconsistent or incomplete results.

\section{Appendix - Potential Enhancements}
\label{ref-appendix-e}

\subsection{Areas of Potential Extension}
\label{ref-030}

Several extensions could broaden Agent Spec’s applicability to common agentic patterns and enterprise-scale requirements while maintaining its declarative and framework-agnostic design principles. Key areas include:
\begin{itemize}
\item \textbf{Memory components:} Representations for short- and long-term memory, including context persistence and retrieval strategies.
\item \textbf{Planning modules:} Declarative definitions of planner components for hierarchical or goal-oriented task decomposition.
\item \textbf{Datastore interfaces:} Native integration with datastores.
\item \textbf{Remote agents:} Support for defining and communicating with remote agents (e.g., using A2A).

\end{itemize}
Each of these enhancements would continue to adhere to the same design principles of clear semantics, explicit I/O, and declarative configuration, while staying framework-agnostic.

\subsection{Runtime Coverage and Conformance}

Expanding runtime coverage remains a core objective. We encourage community-led development of runtime adapters for additional agentic frameworks, along with contributions to an extended conformance test suite. A more robust test suite will help ensure that identical Agent Spec configurations yield consistent behaviors across diverse runtime environments, thereby improving reproducibility, reliability, and interoperability.

\subsection{Improving User Experience}

While the existing SDKs simplify authoring and validating Agent Specifications, future efforts will focus on improving usability and accessibility, particularly for non-programmatic users. Potential directions include:
\begin{itemize}
\item \textbf{Visual authoring tools:} Drag-and-drop interfaces to construct agents and export Agent Spec configurations to be used in the preferred framework via a runtime adapter.
\item \textbf{Advanced diagnostics:} Enhanced schema validation, error explanations, and debugging utilities within SDKs.
\end{itemize}
These efforts aim to make Agent Spec more accessible for practitioners, educators, and enterprise developers alike.

\end{document}